\documentclass[11pt,a4paper]{article}
\usepackage[hyperref]{acl2021}
\usepackage{times}
\usepackage{latexsym}

\usepackage{microtype}

\aclfinalcopy % Uncomment this line for the final submission
\def\aclpaperid{1590} %  Enter the acl Paper ID here

\usepackage{url}
\usepackage{amsmath}
\usepackage{amsfonts}
\usepackage{adjustbox}
\usepackage{verbatim}
\usepackage{enumitem}
\usepackage{graphicx}
\usepackage{xspace}
\usepackage{subcaption}
\usepackage{color}
\usepackage{multirow}
\usepackage{booktabs}
\usepackage{array}
\usepackage{footnote}
\usepackage{xcolor}

\usepackage[size=scriptsize]{todonotes}

\definecolor{pigment}{rgb}{0.2, 0.2, 0.6}
\definecolor{brickred}{rgb}{0.8, 0.25, 0.33}
\newcommand{\consis}{pigment}
\newcommand{\inconsis}{brickred}

\title{Automatic Document Sketching: Generating Drafts from Analogous Texts}

\newcommand\uw{$^{\diamondsuit}$}
\newcommand\ms{$^\spadesuit$}
\newcommand\aspace{\hspace{.75em}}

\author{
Zeqiu Wu\uw\aspace
Michel Galley\ms\aspace
Chris Brockett\ms\aspace
Yizhe Zhang\ms\aspace
Bill Dolan\ms\aspace \\
\uw University of Washington \aspace\ms Microsoft Research\\
{\tt zeqiuwu1@washington.edu}\\
{\tt \{mgalley,chrisbkt,yizzhang,billdol\}@microsoft.com}
}

\date{}

\begin{document}
\maketitle
\begin{abstract}
The advent of large pre-trained language models has made it possible to make high-quality predictions on how to add or change a sentence in a document. However, the high branching factor inherent to text generation impedes the ability of even the strongest language models to offer useful editing suggestions at a more global or {\it document} level.
We introduce a new task, \textsc{document sketching}, which involves generating  entire draft documents for the writer to review and revise. These drafts are built from sets of documents that overlap in form -- sharing large segments of potentially reusable text -- while diverging in content.
To support this task, we introduce a Wikipedia-based dataset of analogous documents 
%(to be released) 
and investigate the application of weakly supervised methods, including use of a transformer-based mixture of experts, together with reinforcement learning. We report experiments using automated and human evaluation methods and discuss relative merits of these models. 
\end{abstract}

\section{Introduction}

\begin{comment}
(1) Large pretrained LM generation (GPT, etc.) can generate long texts that are structurally mostly coherent (while these texts are far from perfect, the global “shape” is usually quite reasonable). Discourse. 
(2) Despite these advances in long-form text generation, most end-user applications for creating texts (e.g., Smart Compose, Smart Reply, Grammarly, Neural Rewrite, Shrimai and Felix’s projects) are still limited to sentence-level generation. 
(3) The challenge in doing multi-sentence generation in a practical setting (i.e., with AI-user interaction) is the high branching factor: user choices of what to include in sentence 1 dramatically affect sentence 2, etc. (butterfly effect).  
(4) Issue in (3) is mitigated with texts that are more structured (boilerplate texts), which constitute a large portion of documents that get written every day (anything to back that up?) 
(5) Motivate the task: document reuse, writer’s block, blank page problem. 
(6) Contributions: new task; new dataset; adaptation of MoE model for that task; address challenge of not having supervised data with RL. Improvement over non neural baseline. 
\end{comment}

Large pre-trained language models such as T5 and GPT-3 \citep{Raffel2019T5, Brown2020gpt3} have enabled impressive progress on a variety of natural language generation tasks by producing fluent, coherent long texts \citep{Rashkin2020PlotMachines, Zellers2019Grover}. 
While automated document-level generation seems tantalizingly within reach, a high branching factor presents significant challenges in tailoring generated documents to the specific requirements of users. 
Topic drift and ``hallucination'' of information are endemic to these models \cite{Wiseman2017ChallengesID}. 
%\cjb{do we have citations on topic drift of GPT-2 generation} 
These risks have ensured that end-user applications involving text generation (e.g., Smart Compose, Smart Reply, Grammarly) still require a human to remain in control of content and are restricted to individual sentences or even smaller segments of text \cite{Chen2016SmartCompose,Kanna2016SmartReply,Alikaniotis2016Grammarly, Prabhumoye2019ContentTransfer, Faltings2020editcommand}.
\begin{figure}
\centering
\includegraphics[width=7.5cm]{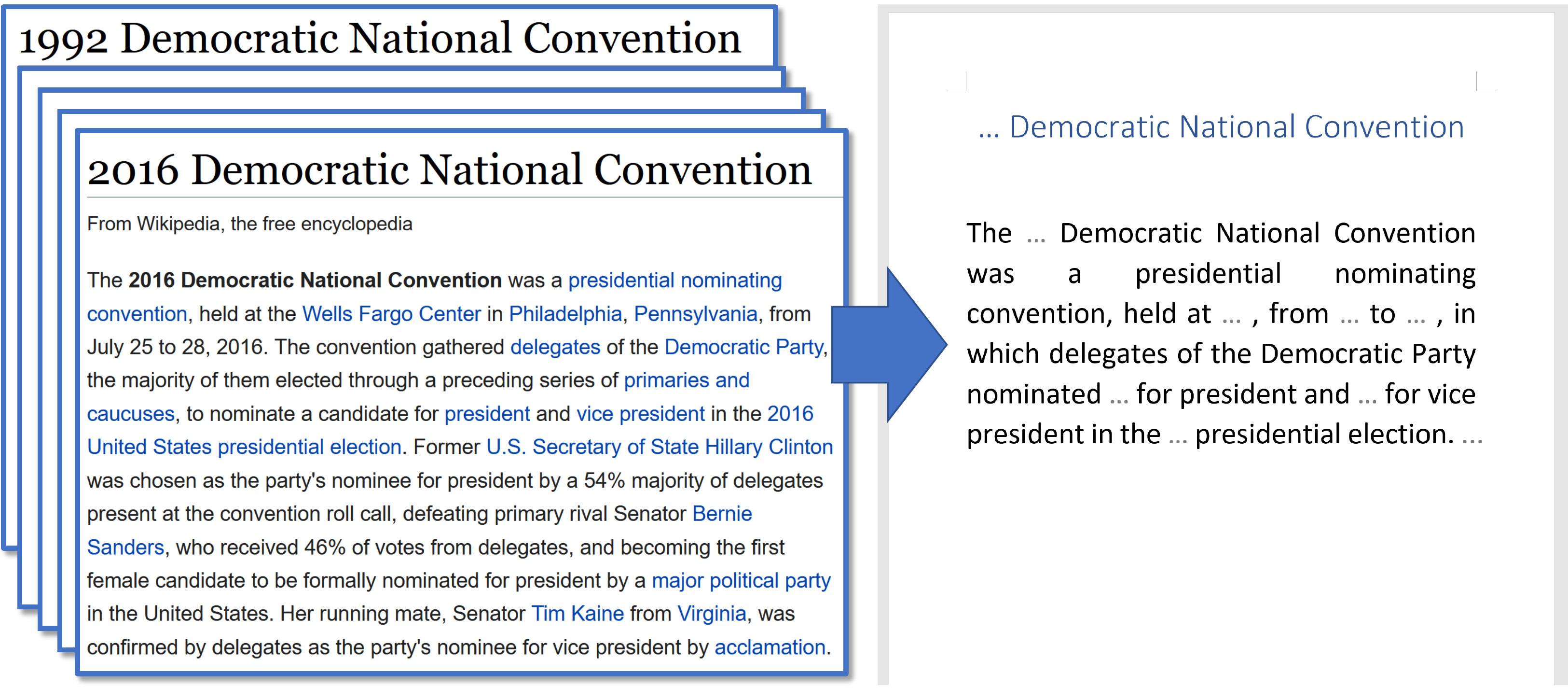}
\caption{\label{fig:teaser} The right side shows a sketch for writing the report of a future democratic national convention, generated from a pile of previous reports. 
}
\end{figure}

Can large generative language models be used to assist user writing at the document level while the user still controls the factual content? 
%create entire new documents while permitting the human to remain in control of content? 
%We begin with the 
A possible answer lies in the
observation that a substantial portion of day-to-day writing involves some form of reuse. Similar documents (e.g., monthly reports, sales letters, job descriptions) are effectively recycled by changing those segments that need to be modified (Fig.~\ref{fig:teaser}).\footnote{Evidenced by the plethora of companies offering reusable business templates for enterprise use, e.g., \url{https://www.businesswire.com/news/home/20201028005573/en/}.}
Moreover, documents containing analogous texts are often found collocated in repositories, a common practice for organizations that manage professional documents.\footnote{For instance, monthly reports of world trade in grains are organized chronologically at \url{https://usda.library.cornell.edu/concern/publications/zs25x844t?locale=en}}
%\footnote{A document management company designed hundreds of. reusable business templates for enterprise use: \url{https://www.businesswire.com/news/home/20201028005573/en/}}. 
%News agencies regularly update their reports on events, and businesses and government agencies maintain large archives that are regularly mined to create new documents.
%\footnote{Evidenced by the plethora of companies offering reusable business templates for enterprise use, e.g., \url{https://www.businesswire.com/news/home/20201028005573/en/}.}
%
%This is not limited to organizations: before writing a reference letter, for example, an advisor might dig out previously written letters with the same purpose to sketch the content.
The high branching factor that impedes the application of long-form generation might thus be mitigated if models were to exploit conventionally-structured analogous ``reusable'' texts to produce a ``sketch'' that captures prototypical document-level text structure. In this view, initial sketches would assist authors by reducing the manual editing effort (planning, inserting and deleting content, etc.). A good sketch would reflect structural patterns and reusable text, and provide indication, beyond static boilerplate, of locations where user input might be warranted. 
This could be especially beneficial for novice writers who would otherwise have to read many analogous documents before developing a full picture of what is entailed in writing such documents.\footnote{Our experiments in Section~\ref{sec:exp} also show that drafts derived from multiple analogous documents are more effective than those derived from one or two documents.}
A fully-implemented dynamic system might update other portions of the document to reflect modifications introduced by the user. 
In this work, we propose a new task, called \textsc{document sketching}, in which initial template-like prototype documents are generated from collections of analogous documents. %based on sample input documents for user editing.
%\yz{"sample input documents" sounds a bit vague}\cjb{I think this is largely solved by moving the desiderata up to the previous paragraph. This para should focus on what we are delivering in this paper.} 
%Such initial sketches should\cjb{the use of "should" sounds  prescriptive} assist human writers by reducing the manual editing efforts (e.g., extra time required for planning, inserting and deleting content, etc). Therefore, a good sketch should contain structural or reusable text, and leave blanks where user input is required. 
To support this task, we collected a dataset consisting of approximately 20K Wikipedia documents with similar textual characteristics.\footnote{We release our data and source code for dataset construction and experiments at \url{https://github.com/ellenmellon/document_sketching}.} 
%
%We define an intuitive automatic evaluation metric based on word error rate to assist system development and 
For this new task, inspired by previous work in measuring machine translation post-editing productivity \citep{Tatsumi2009Postedit, specia2010postedit}, we define an automatic evaluation metric based on Word Error Rate \citep{Snover2006TER,Tomas2003WER}. 
We compare against strong baseline models including a mixture of experts model and a reinforcement learning approach designed to handle multi-source inputs and a weak supervision setting. Finally, we provide experimental analysis of these models, using automated and human evaluation studies.
% MG: not neede do sell paper
%, and outline remaining challenges.

%weakly supervised methods, including use of T5 and T5-based mixture of experts, together with reinforcement learning. We report experiments using automated and human evaluation methods and discuss relative merits of these models.

\section{Related Work}

% Document creation;
\subsection{Document Generation}
Recent work leverages the success of large pretrained language models to generate long texts such as stories \citep{Rashkin2020PlotMachines}, reviews \citep{cho2019towards, shen2019towards} and fake news \citep{Zellers2019Grover}. %,conditioning on predefined meta information. 
Most end-user applications for assisting user writing, however, are  confined to sentence-level generation \cite{Chen2016SmartCompose, Kanna2016SmartReply, Alikaniotis2016Grammarly, Prabhumoye2019ContentTransfer, Faltings2020editcommand}.
% This work 
Our work
focuses on \textit{document-level} writing assistance in which a %n %initial 
document 
sketch is constructed from 
% an  existing 
a set of similar documents. 
%We also encourage future investigation in human-machine interactive writing to provide dynamic writing assistance.

\subsection{Template-Based Generation}
Some existing work induces templates as an intermediate step for performing tasks like text summarization or response generation. Most use a retrieval-based method to extract similar references from the training corpus as prototypes \citep{cao2018templatesum, Yang2019TemplateResponse, Wang2019TemplateSum, Gao2019TemplateSum, Peng2019ExemplarDecode}, and learn to separate salient information and latent template structure. \citet{Cai2019Match2gen} induce an intermediate template for response generation explicitly, but from a single retrieved relevant response. Similar prototype editing work %like 
\cite{Guu2018TemplateSentence,Hashimoto2018StructPred,Fabbri2020TemplateNMT} focuses on short text (e.g., a question or a single sentence) or structured output (e.g., code snippet) editing. \citet{Oya2014TemplateSum,Magooda2019TemplateSum,Yang2020TemplateNMT,Li2018DelRetGen} convert each single input text into a template with blanks using rule-based methods. Other work such as  \cite{Wiseman2018TemplateGen} and
\cite{Gangadharaiah2020TemplateResponse} relies on a %predefined 
knowledge base or a domain/task specific ontology to segment text sequences into templates.

\subsection{Multi-Sequence Processing} 
Multiple Sequence Alignment (MSA), widely used in the biological domain \citep{Sauder2000MSA} to align multiple biological sequences like proteins, has long been leveraged for text pattern matching \citep{Barzilay2003MSA, Alonso2004MSA}. We adopt this method to align input documents and create heuristic templates 
as weak supervision.
%for weakly supervised training. 

Other tasks taking multiple text sequences as input include multi-document summarization, which seeks to generate an abstractive text summary of multiple input documents \citep{Liu2019MultiDocSum, Chu2019MultiDocSum}, and multi-source machine translation \citep{Nishimura2018MultiSourceNMT, Garmash2016MultiSourceNMT} that encodes input texts in multiple source languages and translates them into a target language. \citet{Sang2019MultiDocGen} generate a question from
%based on 
%multiple 
input documents by applying a multi-encoder model with a transformer-based coordinator. 

% Add previous work on MT with translation memory (chris)

\section{Problem Definition}
\label{sec:problem}

We introduce the task of document sketching, which aims to facilitate the authoring process by generating a template-like document draft, based on a collection of sampled similar documents. Formally, the task can be defined as follow: given a set of $n$ documents $X= \{x_1, x_2, ... , x_n\}$, generate a text sequence $s$ that can be used as the sketch to reduce the human effort involved in composing a target document $y$.

\paragraph{Evaluation Metrics}

As in most text generation tasks, we rely on human evaluation (see details in Section~\ref{sec:results}) to draw final conclusions on system comparison. However, due to the high cost of human evaluation, we use automatic evaluation metrics for system development.

It is difficult (and expensive) to collect human-written sketches as references. However, since a generated sketch $s$ is used to help the user complete writing a target document $y$ (i.e., post-editing), we can instead use the target document as a reference and calculate the extra edits required to transform a sketch into that target document. Inspired by the previous uses of word error rate (WER) in evaluating machine translation \citep{Snover2006TER,Tomas2003WER} and evidence of reasonable correlation between WER and human post-editing productivity \citep{Tatsumi2009Postedit, specia2010postedit}, we propose to assess the effectiveness of $s$ in the completion of $y$ based on WER and the Levenshtein distance (abbreviated as `lev') between $s$ and $y$:
\begin{equation} \label{wer_equ}
\begin{split}
  \mathrm{score}(s, y) &= 1-\mathrm{WER}(s, y) \\
              &= 1- \frac{\mathrm{lev}(s, y)}{|y|} 
\end{split}
\end{equation}
The higher $\mathrm{score}(s,y)$ is, the fewer minimum number of word-level insertion and deletion edits a user would need to complete writing $y$ if starting from $s$.
%In order 
To account for the lexical and phrasal variety of writing a target document, we calculate the average score of multiple reference documents $Y$: % instead of a single one: 
\begin{equation} \label{auto_eval_equ}
  \mathrm{score}(s, Y) = \frac{1}{m} \sum_{i=1}^{m} \bigg(1-\frac{\mathrm{lev}(s, y_i)}{|y_i|}\bigg)
\end{equation}
where $Y = \{y_1, y_2, ..., y_m\}$ and $m$ denotes the number of reference documents.

\section{Data}
\label{sec:data}

We collect our dataset from the English Wikipedia dump (June 20, 2020). %Specifically, 
We first group documents into collections %from the Wikipedia dump (of June 20, 2020)
%\footnote{We used the dump snapshot of June 20, 2020.} 
with analogous texts, with the observation that articles with shared reusable structural texts 
%\el{should we provide a better definition of ``reusable text'' as requested by reviewer 3?}\yz{agreed. Better to remind the reader what "reusable" is}
also tend to have similar titles.
%such that each contains documents that are most likely to share reusable texts.
These collections are then split into training, validation and test sets. 
%We describe the details of identifying Wikipedia document collections as well as processing data points below. 
This dataset is designed for a weakly supervised setting, as gold sketches are unavailable.

\subsection{Wikipedia Document Collections}

We observe that Wikipedia article titles provide a strong indication of whether documents are likely to share structural text (i.e., can be put in the same collection) or not. For example, it is reasonable to consider articles with titles like ``Super Bowl~I'', ``Super Bowl~II'' and so on to comprise a document collection of the annual championship game. Therefore, we group documents whose titles are identical but for one token at a specific position. In the ``Super Bowl'' example, the document titles are the same up to the third token in each title, and the collection can be named ``Super Bowl \underline{\hspace{3mm}}''. It is worth noting that collecting these articles based on their titles is simply reflective of how similar documents in Wikipedia tend to cluster. Our task setup only requires a set of references as the input (e.g., documents organized in the same directory), without the need to have reference documents to share titles.

Title matching can yield noise in the extracted document collections, so we apply simple yet effective restrictions in the extraction pipeline. We empirically set the minimum number of documents per collection to 15 and the minimum document title length to 3 tokens. 
%Moreover, for documents of the minimum length 3, we apply an even stronger restriction that the differing token has to be a number. 
We randomly select and inspect over 200 extracted collections, and find 90\% of them contain analogous documents for a certain topic (examples in Tab.~\ref{tab:data_examples}). The remaining 10\% are less clear, but can be usefully grouped into 3 classes as in Tab.~\ref{tab:data_examples}.
To further reduce noise in each collection, we apply
%Furthermore, in order to remove individual documents that are potential noise in each collection, we apply%
Eq.~\eqref{auto_eval_equ} to calculate the average similarity score of each document with the other documents in a collection, removing those with an average score lower than an empirically chosen threshold $-1.5$.
Finally, we keep document collections with more than 5 documents and truncate them to a maximum of 50 documents, which are divided into train, validation, and test sets in the ratio of 0.8/0.1/0.1. The collection statistics are summarized in Tab.~\ref{tab:data}.

\begin{table}
\small
%\begin{adjustbox}{width=0.48\textwidth}
\begin{tabular}{l|p{4cm}}

\toprule
\% Percent (Category) & Example Collection Titles  \\
\midrule
90.0 (reasonable) & \underline{\hspace{3mm}} Academy Awards; Super Bowl  \underline{\hspace{3mm}}  \\
\midrule
10.0 (arguable): &   \\
~~~5.8 (partial name) &  Bob \underline{\hspace{3mm}} (ice hockey) \\
%The Last \underline{\hspace{3mm}} (film);
~~~2.9 (X of a location) & \underline{\hspace{3mm}} of the United States \\
~~~1.4 (other) & Origin of the \underline{\hspace{3mm}} \\
\bottomrule
\end{tabular}
%\end{adjustbox}

\caption{\label{tab:data_examples} Quality analysis of over 200 randomly selected document collections.}
\end{table}

\begin{table}
\centering
\normalsize
\begin{adjustbox}{width=0.48\textwidth}
\begin{tabular}{c  | c c c c }

\toprule
 & Train & Valid & Test & Overall  \\
\midrule
\# Collections & 15.7k & 2.0k & 2.0k & 19.7k  \\
\# Docs / Collection & 21.1 & 21.3 & 21.0 & 21.1  \\
\# Tokens / Doc & 80.6 & 79.3 & 80.3 & 80.5 \\
\bottomrule
\end{tabular}
\end{adjustbox}

\caption{\label{tab:data} Document Collection Statistics.}
\end{table}

\subsection{Task Data Points}
We divide each document collection into multiple smaller collections of analogous documents. For standard supervised training, each data point consists of 1 document $d$ to construct a heuristic sketch as weak supervision (see details in Section~\ref{sec:approach}) and up to 9 input documents $X$. For evaluation, additional 4 documents are used as references $Y$ for calculating the score in Eq.~\eqref{auto_eval_equ}. Since document collections whose title tokens differ by a number usually contain documents about events or entities of different times, we order the documents in ascending numerical order to imitate practical scenarios where humans sketch a document by looking at collections of previously written documents. 
%Since our model relies on reinforcement learning, which also requires such reference documents to calculate the reward for each training data point,
We divide each collection into data points with up to 14 (=1+9+4) documents for all dataset splits, yielding 24k, 3.1k and 3.0k data points (smaller collections) for train, validation, and test sets,  respectively.%\footnote{We will release both extracted document collections as well as processed data points upon acceptance.}

% make a model figure
\begin{figure}
\centering
\includegraphics[width=7.5cm]{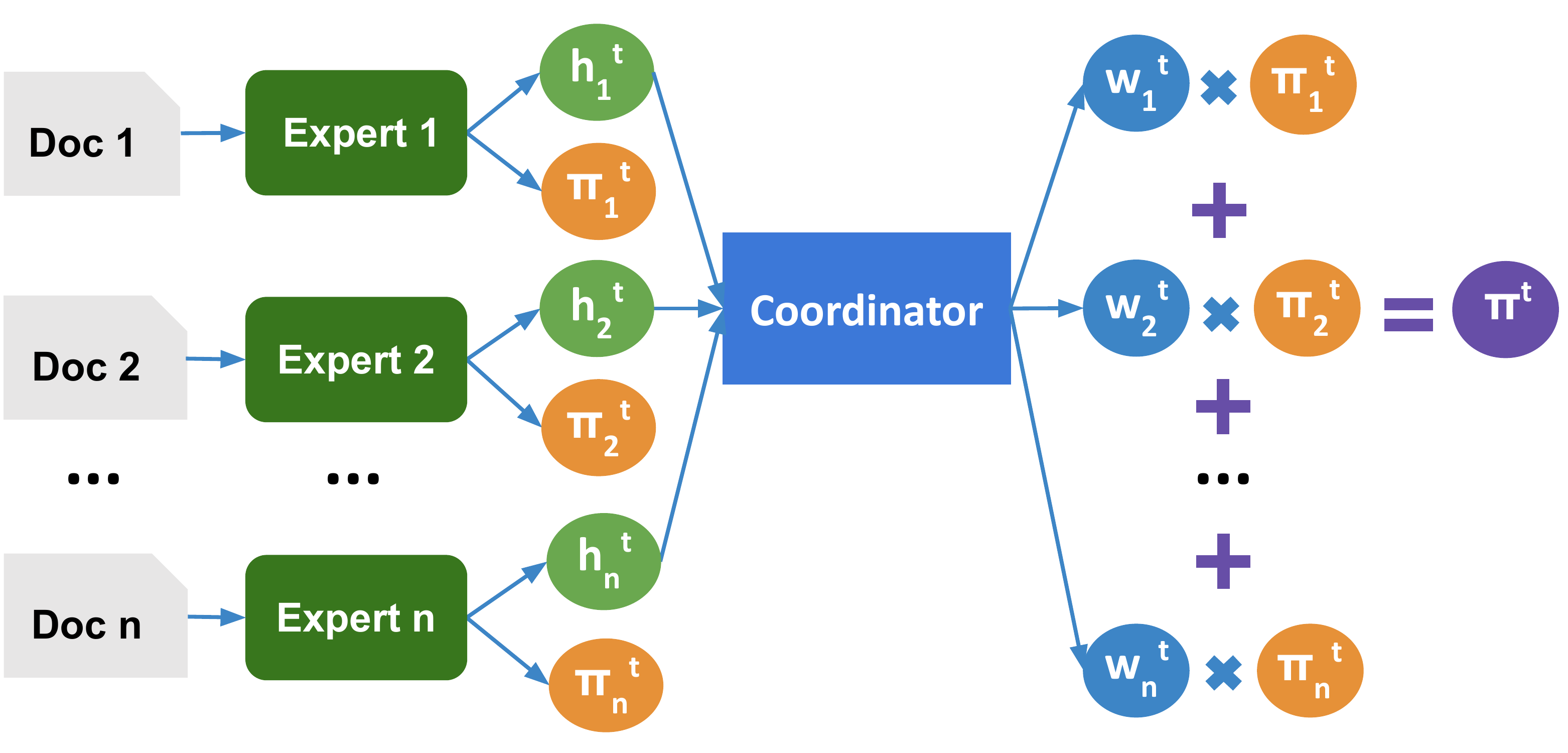}
\caption{\label{fig:moe} Mixture of Experts (MoE) Framework.
}
\end{figure}

\section{Approach}
\label{sec:approach}

Since gold document sketches are not available for supervised training, we first perform weakly supervised training by constructing heuristic sketches as targets. Then we apply reinforcement learning strategies for text generation.

\subsection{Weakly-Supervised Learning}
\label{sec:weak-sl}

\paragraph{Heuristic Labels} As described in Section~\ref{sec:data}, each data point has one document $d$ for creating the weak-supervised sketch. We conduct pair-wise sequence alignment
(using the Ratcliff-Obershelp algorithm \cite{Ratcliff&Metzener:88})
%% If it's a standard algorithm: either cite an old paper, textbook, or nothing.
%\footnote{We use the Python class difflib.SequenceMatcher for sequence alignment.} 
for $d$ and each input document $x_i \in X$, and count the relative alignment ratio (the number of tokens being aligned divided by $|X|=n$) of each token in $d$. We retain tokens with this ratio above a threshold and replace other tokens with an ellipsis token to form the heuristic target sketch $s$ given $X$. 
Consecutive ellipses are merged.
The threshold is empirically set at 0.6, which yields the highest average $\mathrm{score}(s, Y)$ on the validation set.

%\paragraph{T5}
%The T5 model \citep{Raffel2019T5} is an encoder-decoder architecture pre-trained on a variety of text-to-text tasks. To leverage the recent success in such transformer-based generation models, 
%% maybe too defensive?:
%as well as to draw fair conclusions from system comparisons, 
%all sequence-to-sequence (S2S) generation models in our experiments are initialized with T5-base.

\paragraph{Mixture of Experts (MoE)}
To generate an initial document sketch, the model needs to take multiple input documents. The most obvious way to do this is to concatenate all input documents into a single long sequence and feed that into an encoder-decoder model like T5 \citep{Raffel2019T5}. However, processing long sequences in such models is memory-consuming and lack of structure makes it difficult for the model to perform document-level coordination.
We therefore use a mixture of experts (MoE) framework, in which a coordinator decodes a token at one timestamp by taking the hidden state and the output vocabulary distribution from each expert that processes a single document. Fig.~\ref{fig:moe} shows the overall framework of MoE. 

The experts have the same encoder-decoder structure and aim to generate $s$ by each encoding a separate single input document only. All experts share the same model parameters. At each decoding timestamp $t$, the $i^{th}$ expert encodes $x_i$ as well as previously decoded tokens from a coordinator $\Tilde{s}_{1}...\Tilde{s}_{t-1}$ (or $s_{1}...s_{t-1}$ during training), and outputs a probability distribution $\pi_i^t$ over all vocabulary words. The coordinator is a transformer-based encoder that takes the hidden state at the current timestamp $h_i^t$ of each $i^{th}$ expert, and outputs a weight $w_i^t$ with a final linear layer. The output weights are used to calculate a weighted sum of the probability distributions from their corresponding experts $\pi_t = \sum_{i=1}^{n} w_i^t \, \pi_i^t$, where $\pi_t$ is the final distribution 
%used to do 
for
generation at timestamp $t$.

\subsection{Reinforcement Learning}
We leverage reinforcement learning (RL) 
%following the implementation from \citet{zhang2019dialogpt} 
to further improve generation quality. For each training example with input $X$, we generate a sequence $\hat{s}$, which is sampled from the probability distribution at each time step, $p(\hat{s}_t|\hat{s}_1...\hat{s}_{t-1}, X)$.
We observe that directly optimizing the evaluation function proposed in Eq.~\eqref{auto_eval_equ} at the sequence level, using a vanilla policy gradient (PG) or a self-critical sequence training (TD-SCST) algorithm \citep{rennie2017self-critical, Paulus2018RLSum, Pasunuru2017RLSum}, can lead to instability during training as the reward cannot be calculated until the end of generation \citep{Celikyilmaz2018DeepComAgent}. 
Therefore, we instead use a token-level incremental reward that is based on the change to the original reward function $r(\hat{s}, Y)=\mathrm{score}(\hat{s}, Y)$ from each sampled token $\hat{s}_t$, given references $Y$: 
\begin{equation}
r_t(\hat{s}_t, Y) = r(\hat{s}_{1...t}, Y) - r(\hat{s}_{1...t-1}, Y).
\end{equation}
The training objective can be written as:
\begin{equation}
  L_{\mathrm{RL}} = \sum_{t=1}^{T} - r_t(\hat{s_t}, Y) p(\hat{s}_t|\hat{s}_1 . . . \hat{s}_{t-1}, X)
\end{equation}
where $T=|\hat{s}|$. Since optimizing RL loss alone runs the risk of compromising the language model \citep{Paulus2018RLSum, Pasunuru2017RLVideoCaption}, we use a mixed loss as follows:
\begin{equation}
    L_{\mathrm{MIX}} = \lambda L_{\mathrm{RL}} + (1-\lambda) L_{\mathrm{MLE}}
\end{equation}
where $\lambda$ is a hyperparameter to be tuned.

\section{Experiments}
\label{sec:exp}

\subsection{Setup}
%\subsection{Implementation Details}
% parameters, base T5 model, etc
% infrastructure, # times being tuned, etc
To leverage the recent success in such transformer-based generation models, 
neural generation models in our experiments are initialized with the base version of T5 \citep{Raffel2019T5,wolf2020huggingfaces}, an encoder-decoder architecture pre-trained on a variety of text-to-text tasks.\footnote{Note that T5 can be easily replaced by Other pre-trained generative models like BART \citep{lewis2020bart} in our model framework.}
%\footnote{\url{https://huggingface.co/transformers/model\_doc/t5.html}} 
All hyperparameters are tuned on the validation set. For MoE, we first fine-tune T5-base to obtain a ``single expert'' model to initialize each individual component model of the MoE. The ``single expert'' is trained by leveraging a sequence-to-sequence objective as in T5, with each of the $n$ training (input, output) pairs: $(x_1, s)$, $(x_2, s)$, ..., $(x_n, s)$ from a data instance. 
This is followed by training a complete MoE model as described in Section~\ref{sec:approach}. We use greedy beam search as the decoding strategy for all generative models with a beam size of 4 (the default value in T5-base).
% post-processing
We observe that consecutive ellipses and uninformative tokens hurt readability. To improve the readability of output sketches, we apply minor post-processing to all models in our experiments: we merge consecutive ellipses if all tokens between them are punctuation or among the 30 most frequent tokens.\footnote{\url{https://github.com/first20hours/google-10000-english}} Sentence-terminating periods are excepted. %The only exception is the period mark which we exclude owing to its importance as sentence terminator. 

\paragraph{Training and Parameters} T5 has about 220 million parameters and MoE has about 310 million parameters in total. 
% average run time
The average training time is about 20 hours for T5 and about 30 hours for MoE on a single Tesla V100 node (32GB) with 3 epochs. It takes an additional 10 hours to train a single expert in MoE. For MoE+RL, we use 4 V100 nodes and it takes about 2 days for training.
% bounds for each hyperparameter, times of training
% The method of choosing hyperparameter values (e.g. manual tuning, uniform sampling, etc.) and the criterion used to select among them (e.g. accuracy)
During training, we tune batch size, learning rate and warm-up steps for T5. For MoE and MoE+RL, there is a separate pair of learning rate and warm-up steps for the coordinator. Each batch size is tuned within a range of [1, 32] (with no more than 3 values for each model); each warm-up step is tuned at $\{4000, 8000, 16000\}$; each learning rate is tuned in $\{1e^{-5}, 3e^{-5}, 1e^{-4}\}$. For RL training, we observe performance gain only when $\lambda$ is relatively large, so we tune $\lambda\in\{0.95, 0.97, 0.99\}$ (similar to \citet{Celikyilmaz2018DeepComAgent}). In order to avoid the situation where the models learn to not generate ellipses that lead to unreadability, we set the cost for deleting an ellipsis to be much smaller (0.1) in the reward function. In addition, we randomly select three seed values for each model. Each model is trained with the above parameter value combinations. We apply grid parameter search for T5 and MoE, and random search for MoE+RL. Hyperparameters are tuned based on automatic evaluation score (1-WER) on the validation set.

% hyperparameter configurations for best-performing models 
The best hyperparameter configurations for best-performing models were: \textbf{T5}: batch size = 2, warm-up step = 4000, learning rate = $3e^{-5}$; \textbf{MoE}: batch size = 15, warm-up step = 8000, learning rate = $3e^{-5}$, warm-up step (coordinator) = 4000, learning rate (coordinator) = $1e^{-4}$; \textbf{MoE+RL}: batch size = 4, warm-up step = 4000, learning rate = $3e^{-5}$, warm-up step (coordinator) = 4000, learning rate (coordinator) = $3e^{-5}$, $\lambda=0.99$.

\subsection{Systems}
%\subsection{Compared Systems}
\label{sec:systems}

%We provide results and comparisons for a variety of both neural and non-neural models. 

Non-neural systems include the following approaches: \textbf{Last-pair}: The output is the aligned boilerplate text between the last pair of ordered input documents (as described in Section~\ref{sec:data}, in cases where the differing title token is time-related, the last pair refers to the most recent two documents); \textbf{Last-retrieval}: The last document is retrieved as the initial draft; \textbf{MSA}: We apply the
multi-sequence alignment approach \citep{Barzilay2003MSA}, to align tokens in input documents and replace tokens that have too few alignments (threshold tuned on valid set) with ellipses;
\textbf{Consensus-MSA}: This resembles how we generate heuristic drafts as weak supervision in Section~\ref{sec:weak-sl}. However, instead of using the held-out document $d$ to create the skeleton, we use the input document $x_i$ that gives the highest value of $\sum_{j=1}^{n} \mathrm{score}(x_i, x_j)$.

%\paragraph{Neural} 
We evaluate the following neural approaches:
\textbf{T5 (zero-shot)}: We directly apply T5 without finetuning. We use T5 to encode each input document individually and at each decoding timestamp, we combine probability distributions from all T5 decoders with average pooling. 
%The best token from the combined distribution will be decoded if its probability is above a threshold (tuned on valid set), otherwise an ellipsis token will be decoded.
As T5 rarely generates ellipses, we insert an ellipsis token if the probability of the top candidate token $w$ from the combined distribution is below a threshold $\alpha$ that is tuned on the validation set (i.e., $p(w)<\alpha$).\footnote{Since the original T5 
was not trained with ellipsis as a special token,
%may not know how to handle ellipses, 
%we made each individual 
each T5 decoder uses its own greedily decoded token if an ellipsis is inserted ($p(w)<\alpha$).} We used the ``summarize~:'' prefix in the zero-shot setting, this being is the most similar  T5 pre-trained tasks to ours; \textbf{T5 (doc-finetune)}: We finetune T5 with all input documents concatenated into a single string input (with a special separator token) and each target document (instead of the heuristic sketch) as the generation target; \textbf{T5}: Similar to the \textit{T5 (doc-finetune)} setting, but with the heuristic sketch defined in Section~\ref{sec:approach} as the generation target; \textbf{MoE}: This is the mixture of experts model described in Section~\ref{sec:approach}, also trained with the heuristic sketches as the weak supervision; \textbf{MoE + RL}: This is the RL model described in Section~\ref{sec:approach}, with the trained MoE as the warm-start. 

\subsection{Results}
\label{sec:results}
%\subsection{Quantitative Results}

\begin{table}
\centering
\small
%\begin{adjustbox}{width=0.48\textwidth}
\begin{tabular}{l@{}r|rrr}

\toprule
%\multirow{3}{*}{} & \multicolumn{1}{c|}{} & \multicolumn{3}{c}{}\\
&& \multicolumn{3}{c}{Input document similarity}\\
System & All & High & Med. & Low \\
%Input Doc Coherence & All & High & Medium & Low \\
\midrule
last-retrieval &$-$0.242 & 0.338 & $-$0.284 & $-$0.733 \\
last-pair& 0.147 & 0.410 &  0.063 & $-$0.021 \\
MSA& 0.149 & 0.374 &  0.072 & 0.012 \\
consensus-MSA& 0.202 & \bf{0.438} &  0.132 & 0.047 \\
\midrule
T5 (zero-shot) & 0.000 & 0.000 & 0.000 & 0.000 \\
T5 (doc-finetune) & 0.050 & 0.382 & $-$0.014 & $-$0.197 \\
T5 & 0.182 & 0.389 &  0.124 & 0.044 \\
MoE & 0.205 & 0.411 &  0.147 & 0.069 \\
% MoE+RL& 0.211 & 0.410 &  \bf{0.159} & 0.073 \\
MoE+RL& \bf{0.213} & 0.414 &  \bf{0.158} & \bf{0.077} \\
\bottomrule
\end{tabular}
%\end{adjustbox}

\caption{\label{tab:auto} Automatic evaluation scores (1$-$WER) of overall test examples and scores categorized by input document similarity levels.}
\end{table}

\begin{table}
\centering
\small
%\begin{adjustbox}{width=0.48\textwidth}
\begin{tabular}{ll|rr}

\toprule
%\multirow{3}{*}{} & \multicolumn{1}{c|}{} & \multicolumn{3}{c}{}\\
%Model A & Model B & Prefer A & Prefer B \\
System A & System B & Prefer A & Prefer B \\
\midrule
MoE & consensus-MSA & \bf{50.70\%} & 39.28\% \\
MoE & T5 &  \bf{46.15\%}  & 42.88\% \\
MoE & MoE+RL & \bf{49.93\%} & 39.65\% \\
\bottomrule
\end{tabular}
%\end{adjustbox}

\caption{\label{tab:human} Human evaluation scores. A number in bold indicates that the system is significantly better at \mbox{$p < 0.03$}, computed using 10k bootstrap replications.}
\end{table}
\begin{table*}[h]
\centering
{\small
%\begin{adjustbox}{width=1.0\textwidth}

\begin{tabular}{c|p{12.9cm}}

\toprule
Collection Title & \underline{\hspace{3mm}} Waterford Senior Hurling Championship \\
Target Document & The `1960 WSHC' was the 60th staging of the WSHC since its establishment by the Waterford County Board in 1897. On 9 October 1960, Mount Sion won the championship after a 5-09 to 2-05 defeat of Erin's Own in the final. This was their 16th championship title overall and their eighth title in succession. \\
\midrule
consensus-MSA & \textcolor{\consis}{The `... WSHC' was the} ... \textcolor{\consis}{staging of the WSHC since its establishment by the Waterford County Board in 1897. On} ...\textcolor{\consis}{, Mount Sion won the championship after a} ... \textcolor{\consis}{defeat of} ... \textcolor{\consis}{in the final. This was their} ... \textcolor{\consis}{championship title overall and their} ... \textcolor{\consis}{title in succession.} \\
\midrule
T5 & \textcolor{\consis}{The `}... \textcolor{\consis}{WSHC' was the} \textcolor{\inconsis}{58th} \textcolor{\consis}{staging of the WSHC since its establishment by the Waterford County Board in 1897. On} ... \textcolor{\inconsis}{September} ... \textcolor{\consis}{won the championship after a} ...\textcolor{\inconsis}{-10} \textcolor{\consis}{to} \textcolor{\inconsis}{1-}... \textcolor{\consis}{defeat of} ... \textcolor{\consis}{in the final. This was their} ... \textcolor{\consis}{championship title overall and their} ... \textcolor{\consis}{title in succession.} \\
\midrule
MoE & \textcolor{\consis}{The `}\textcolor{\inconsis}{1957} \textcolor{\consis}{WSHC' was the} \textcolor{\inconsis}{57th} \textcolor{\consis}{staging of the WSHC since its establishment by the Waterford County Board in 1897. On} \textcolor{\inconsis}{15 September 1957}\textcolor{\consis}{, Mount Sion won the championship after a} \textcolor{\inconsis}{2-10} \textcolor{\consis}{to} \textcolor{\inconsis}{1-08} \textcolor{\consis}{defeat of} ... \textcolor{\consis}{in the final. This was their} \textcolor{\inconsis}{13th} \textcolor{\consis}{championship title overall and their} \textcolor{\inconsis}{fifth} \textcolor{\consis}{title in succession.} \\
\midrule
MoE+RL & \textcolor{\consis}{The `}... \textcolor{\consis}{WSHC' was the} ... \textcolor{\consis}{staging of the the WSHC since its establishment by the} ... \textcolor{\consis}{County Board in 1897. On} ... \textcolor{\inconsis}{1957,} ... \textcolor{\consis}{won the championship after a} ...\textcolor{\inconsis}{-10} \textcolor{\consis}{to} ... \textcolor{\consis}{defeat of} ... \textcolor{\consis}{in the final. This was their} ... \textcolor{\consis}{championship title overall and their ... title in succession.} \\
\midrule\midrule
Collection Title & Botanischer Garten der Universität \underline{\hspace{3mm}} \\
Target Document & The `Botanical Garden in Potsdam' (or Botanischer Garten der Universität Potsdam), is a botanical garden and arboretum maintained by the University of Potsdam. It has a total area of 8.5 hectares, of which 5 hectares are open to the public, and is located immediately southwest of the Orangery Palace at Maulbeerallee 2, Potsdam, in the German state of Brandenburg. It is open daily; an admission fee is charged for the glasshouses only (2017). The garden was established in 1950 on two adjacent plots of land: part of the Sanssouci Park, and the Paradise Garden (about 2.5 hectares).\\
\midrule
consensus-MSA & ... \textcolor{\consis}{a botanical garden} ... \textcolor{\consis}{maintained by the University of} ... \textcolor{\consis}{It} \textcolor{\inconsis}{is located} ... \textcolor{\consis}{open} ... \textcolor{\consis}{The garden was} ... \textcolor{\inconsis}{contains} ... \\
\midrule
T5 & \textcolor{\consis}{The 'Botanischer Garten der Universität} ... \textcolor{\consis}{is a botanical garden maintained by the University of} ... \textcolor{\consis}{ It is located at} ... \textcolor{\inconsis}{Baden - Württemberg, Germany, and} \textcolor{\consis}{open} ... \textcolor{\consis}{The
garden was established in} ... \textcolor{\inconsis}{and contains about} ... \textcolor{\inconsis}{species} ... \\
\midrule
MoE \& MoE+RL & \textcolor{\consis}{The 'Botanischer Garten der Universität} ... \textcolor{\consis}{is a botanical garden maintained by the University of} ... \textcolor{\consis}{It is located at} ... \textcolor{\consis}{The garden was established in} ... \\
%\midrule
%MoE+RL & \textcolor{\consis}{The `}... \textcolor{\consis}{United States House of Representatives elections' was held on November} ... \textcolor{\consis}{.} \\
\bottomrule

\end{tabular}
%\end{adjustbox}
\caption{\label{tab:case-study} 
Sample outputs:
%Two sample examples of system outputs. 
in the first set of outputs there is high similarity among documents in the input document set while in the second inter-document similarity is low. Generated tokens are in \textcolor{\consis}{blue} if they are consistent with the target document, and in \textcolor{\inconsis}{red} if inconsistent.
%the first has high similarity among documents in the input document set while the second has low similarity. Generated tokens are in \textcolor{\consis}{blue} if consistent with the target document, and in \textcolor{\inconsis}{red} if inconsistent.
}
}
\end{table*}

\paragraph{Automatic Evaluation} 
%As explained in Section~\ref{sec:problem}, we use automatic evaluation 
We use the automatic evaluation of Section~\ref{sec:problem} 
for system development while relying on human evaluation to draw final conclusions on system comparison. However, we observe reasonable consistency between our automatic and human evaluation results. We report automatic evaluation results in Tab.~\ref{tab:auto}. The first numerical column includes overall results of the test set, and the remaining columns categorize the results into three roughly equi-sized levels of similarity among documents within the input document set, estimated on the basis of average pair-wise edit distance normalized by length.

% compare msa with other non-neural baselines
Among non-neural models, we observe that multi-sequence alignment approaches outperform last-retrieval and last-pair where only one or two input documents are leveraged to produce the initial draft. We also notice that consensus-MSA with a ``central'' document as the skeleton is more effective than the standard MSA that treats all inputs equally.
% briefly talk about T5 zero shot
When directly applying zero-shot T5, we notice that when the threshold $\alpha$ described in Section~\ref{sec:systems} equals 1.0 (i.e., the output is always an empty string), the model can achieve the best score 0. This is partly because that none of the T5 pre-trained tasks is the same as ours. In addition, simply exploiting the generation probability at each decoding timestamp can be problematic. For example, when decoding a frequently-appearing entity with multiple tokens, the first token might have low probability given the preceding context while subsequent tokens can be more probable simply because they frequently appear with the first token.
% compare using target document as label and heuristic draft as label
T5 with supervision from the target document (doc-finetune) yields a much lower score than T5 finetuned with the heuristic draft. This is mostly because the complete target document contains much more information than can be inferred from the input; the model learns to hallucinate facts, necessitating heavy deletion by users.
%due to information contained in the target document cannot be inferred from the input; the model learns to hallucinate facts with such training supervision.

% compare t5 and moe
MoE outperforms T5, which indicates that having a coordinator to communicate between each individual encoder and decoder effectively improves model performance. 
% MoE coordinator : linear or transformer-based
We also experiment with coordinators using a simple linear layer and find that transformer-based ones are much more effective.
% compare moe and rl
Applying RL on top of a trained MoE model helps further improve the model performance in automatic evaluation. We denote the RL approach as MoE+RL in Tab.~\ref{tab:auto}. 
% RL (better and worse starts)
Warm-starting RL with a MoE model close to fully-converged
%% Trying to cut space, can put this back for camera-ready:
%\footnote{Fully-convergence is indicated by achieving best performance on development set.} 
gives slightly better results than with an fully-converged MoE model.

% mention different similarity.
The rightmost three columns in Tab.~\ref{tab:auto} divide test examples into 3 roughly equal-sized levels of similarity among documents within the input document set (`high', `medium', `low'). Both last-retrieval and T5 (doc-finetune) models drop dramatically as the input similarity decreases, mostly because they contain a lot of hallucinated content, an issue that is even more severe when there is little overlapping structure or factual content among the input documents. MoE based models are much more robust to low input similarity compared to the baselines. For the group with the most similar input documents, consensus-MSA gives the best score, while MoE based models yield much better performance in the other two groups.

\paragraph{Human Evaluation}
% explain that we controls length and % dots in all system outputs
In order to avoid bias in human judgments, we control possible confounding factors including sequence length and the number of ellipses in a sequence during decoding  \citep{Nakov2012MTLength, Guzman2015MTLength}. We apply a tuneable penalty for each confounding factor (at inference time only) for each neural model \citep{Murray2018MTLengthBias} to generate examples for human evaluation, such that the average output length and number of ellipses in each sequence from all neural systems compared in Tab.~\ref{tab:human} are almost the same.\footnote{Since MoE has a more balanced average sequence length and the number of ellipses, we keep MoE unchanged and adjust T5 and MoE+RL to have the same values as MoE.} 
We notice that such normalization does not affect the automatic evaluation score of each system much and the system ranks do not change.

% report human eval results
Human evaluation was conducted using crowd-sourced workers. 
Judges were presented with paired randomized outputs and target documents, and were instructed to choose their preference for a
%on which output as the 
starting point for writing the target document in order to 
save editing time.\footnote{
%save time and editing effort.
Ideally, it would be preferable to directly measure 
editing productivity through usability testing, but this is impractical in a crowd worker setting, and dependent on confounding factors such as design of the user interface.
%the time spent by human writers on completing the target document 
%when given a system generated sketch is 
%would be
%the most ideal setup for our proposed problem setting, it is practically very hard to implement with crowd workers.
}
Judgments were based on a five-point Likert scale, and ties were permitted. Four judges evaluated each pair, and metrics were imposed to block poorly performing judges. 
%Inter-rater agreement was XXX with Krippendorff's coefficient at XXX.
Sample sizes are 1000 for all system pair comparisons.

\begin{table*}%[h]
\centering
{\small
%\begin{adjustbox}{width=1.0\textwidth}
\begin{tabular}{c|p{12.9cm}}

\toprule
Collection Title & 1960 United States Presidential Election in \underline{\hspace{3mm}}\\
Target Doc Title & 1960 United States Presidential Election in Colorado\\
%User Input & The '1960 United States presidential election in Colorado' took place on November 8, 1960, as part of the 1960 United States presidential election.\\
\midrule
w/o user input & \textbf{The `1960 United States presidential election in ...' took place on November 8, 1960, as part of the 1960 ... election.} ... chose ... representatives, or electors to the Electoral College, who voted for president and vice president.\\
w/ user input & \textbf{The `1960 United States presidential election in Colorado' took place on November 8, 1960, as part of the 1960 United States presidential election.} \textcolor{\consis}{Voters} chose \textcolor{\inconsis}{three} representatives, or electors to the Electoral College, who voted for president and vice president. \textcolor{\consis}{Colorado was won by incumbent Vice President Richard Nixon (R-California), with} \textcolor{\inconsis}{50.9\%} \textcolor{\consis}{of the popular vote, against Senator John F. Kennedy (D-Massachusetts) with} \textcolor{\inconsis}{49.1\%.}\\
\bottomrule
\end{tabular}
%\end{adjustbox}
\caption{\label{tab:interactive} Example of interactive writing with user input. The first sentence (\textbf{bold}) is generated in the ``w/o user input'' setting, while it is the gold first sentence of the target document in ``w/ user input'' setting. 
%Extra 
%Generated tokens in the following sentences under the user input setting are in  \textcolor{\consis}{blue} if consistent with the target document and \textcolor{\inconsis}{red} if inconsistent.
Generated tokens 
%% Can we drop the following:
%in the following sentences under the user input setting 
are in  \textcolor{\consis}{blue} if consistent with the target document and \textcolor{\inconsis}{red} if inconsistent.
}
%\textcolor{\consis}{blue} (consistent with target document) or %\textcolor{\inconsis}{red} (inconsistent).
%}
}
\end{table*}
\begin{table}
\centering
\small
%\begin{adjustbox}{width=0.48\textwidth}
\begin{tabular}{l|rr|rr}

\toprule
\multirow{3}{*}{} & \multicolumn{2}{c|}{Ellipses} & \multicolumn{2}{c}{Punctuation}\\
System & High & Low & High & Low \\

\midrule
MSA& 15.9\% & 42.0\% & 10.9\% & 19.0\% \\
consensus-MSA& 10.6\% & 29.8\% & 10.8\% &  19.2\% \\
T5 & 10.0\% &  23.2\% & 10.9\% & 14.6\% \\
MoE & 8.6\% &  16.6\% & 10.2\% & 13.0\% \\
MoE+RL& 7.3\% &  15.4\% & 10.2\% & 14.0\% \\
\bottomrule
\end{tabular}
%\end{adjustbox}
\caption{\label{tab:analysis} Percentage of ellipses (in all generated tokens) and punctuation (in generated tokens excluding ellipses) tokens. `High' and `low' refer to the average pair-wise input document similarity level.}
\end{table}

% draw conclusions
Results in Tab.~\ref{tab:human} confirm that MoE outperforms consensus-MSA and T5. Although RL helps improve automatic evaluation scores on top of MoE, human judges mostly prefer MoE over MoE+RL.  
% explain why RL didn't win in human eval
The fact that applying RL can hurt readability \citep{Paulus2018RLSum, Pasunuru2017RLSum} may explain why MoE+RL achieves higher automatic scores yet worse human scores than MoE. We observe that MoE+RL has occasional difficulty predicting where ellipsis tokens should be, which hurts sketch readability. Moreover, since our task is a subjective one (e.g., different preferences for more/less verbose sketches), a customized RL reward (e.g., different cost for deletion and insertion for WER calculation) should be applied in real applications to reflect different user preferences. 

%{\bf TODO}: add human eval on visualization.

%\subsection{Qualitative Analysis}
\subsection{Analysis}

We investigated the differences in model performance when the input documents have different similarity scores. When the similarity is lower, models tend to produce drafts that are harder to  interpret because they generate a higher percentage of generic or ellipsis tokens, although MoE and MoE+RL are less vulnerable to this. Tab.~\ref{tab:analysis} shows the average percentage of generated ellipsis or punctuation tokens for each system when the input document similarity is high or low. We can see that when the similarity score is higher, the two statistical numbers do not differ greatly between systems, except that MSA has a much higher percentage of ellipses. When similarity is low, MoE and MoE+RL have much lower percentages of ellipses or punctuation, compared with other models.

% examples of different models

Tab.~\ref{tab:case-study} shows two examples of system outputs.
%\mg{Seeing a page of a scientific paper with font predominantly in blue is a bit odd. I understand the reason here, but black should be the default. What about blue to black, red to grey? I suggest using macros so we don't have to hard code the color at every place.}.
These examples are from the clusters of high and low input similarity respectively. When the input similarity is high, content tokens from consensus-MSA are included in the target document while the neural models tend to generate more hallucinated tokens. This explains the better score consensus-MSA achieves in the higher input similarity cluster. We also notice that when the similarity is high, consensus-MSA generally has better coverage of overlapped content between input documents, and its generated drafts are on average more than 10\% longer than other systems. When the similarity is low, the second example shows that all systems generate shorter outputs due to less shared structure between input documents. In this case, consensus-MSA starts to include more content that does not necessarily appear in the target and MoE-based models generate more reasonable sketches and contain fewer ellipses and uninformative tokens, which can hurt the sketch readability.

% e.g. different costs for different correlation. different result table for the best cost set.
%Our WER-based automatic evaluation metric has a fair correlation with human evaluation, with a XXX Spearman's correlation and XXX p-value. However, with the observation that too many filler tokens may hurt the draft readability as well as that personalized writing habits may lead to different preference over provided drafts (e.g. token insertions are more preferred than deletions, or vice versa), we calculate our WER-based score differently by assigning different costs to insertion as well as deletion of each content or filler token. We get the best cost combination over all human evaluated examples to be: XXX, which gives a higher Spearman's correlation: XXX. We also notice that the best cost combination that gives the highest correlation is sensitive to different system comparisons. Therefore, although we provide the newly calculated scores for each system in Table XXX with the best cost combination, we stick with the standard costs where every cost is 1.0. \el{add tables, numbers, etc} 
% For the final NAACL submission, add one paragraph for using RL to optimize towards the best cost set

%\paragraph{Discussion}
\section{Discussion}

Since gold document sketches are difficult to annotate, there remain challenges of how to leverage or create better weak supervision. 
%Additionally, we may 
We may also
want models to adjust to users' personalized preferences over document sketches (e.g., some may prefer more deletions than insertions or vice versa) in practical application deployment, where RL
%reinforcement learning 
can play a major role by having models directly optimized by customized rewards.

We show that MoE-based models tend to generate more readable sketches by outputting fewer ellipses and functional tokens, though in some cases it can be difficult to guess what should fill a given ellipsis.
%but there are some cases that are hard to interpret what to fill in each ellipsis. 
An interesting future study could examine how the placement and number of ellipses in a sketch affects readability. We would also like to explore whether replacing ellipsis tokens with some more contentful label (e.g., entity type)
%or semantic topic)
or a retrieved sample content for a gap \citep{xu2021ide} could help make sketches more 
%readily 
interpretable.

%Fully automated document generation is challenging, partly due to the high branching factor that each local semantics can be very sensitive to users' choices of its surrounding context. However, a
An initial document sketch provides the reusable text structure that allows a user to fill in detailed content. Interactive document generation can be seen as the next step of our document sketching task that dynamically fills in more local contexts based on users' inputs. We suggest that RL could again play a major role towards building such writing assistants, with rewards from a real user or a user simulator. %We show examples from a very simple user simulator that assumes the user modifies the initial draft to be exactly the same as the target document in a left-to-right order. 
%We use the trained MoE model to encode the input documents and force the decoder to start with the first sentence in the target document. 
%In the example from Tab.~\ref{tab:interactive}, 
We notice, for example, in Tab.~\ref{tab:interactive} that given user input in the first sentence, the subsequent generated sentences  become much more specific and relevant to the new input.

\section{Conclusions}

We have presented a new task, \textsc{Document Sketching}, designed to generate draft documents that users can edit.  
To support this task, we introduced a new dataset 
%extracted from Wikipedia, 
and a weakly supervised learning setting. Experimental results show that deep learning models outperform established multi-sequence alignment approaches. 
%It is our hope that some future implementation will eventually help unblock writers 
%facing a blank page. 
%who find themselves facing a blank page. 

\section*{Acknowledgments}
We thank members of Microsoft Research and University of Washington's NLP groups who provided feedback and insights to this work. In particular, we thank Chenyan Xiong for providing insights during early project discussions. We also thank Michael Lee, Sara Ng, Sitong Zhou, and Trang Tran for their assistance and feedback on the human evaluation setup.
We also thank Zhang Li, Kosh Narayanan, Chandra Chikkareddy, Si-Qing Chen, and Weixin Cai of  Microsoft's Natural Language Experience team for their insights into authoring experience.

\section*{Ethical Considerations}

We believe this paper to be part of body of work that can mitigate some of the ethical pitfalls \cite{bender:21} of large pre-trained models such as GPT-3 \cite{Raffel2019T5}.
While the present work does not entirely prevent misuse, e.g., by a malicious user trying to promote misinformation, Automatic Document Sketching can be more broadly framed as a form of {\it controllable} generation, which places greater control into the hands of the users. %While this work does not entirely prevent misuse (e.g., by a malicious user trying to promote misinformation), %we believe this work 
As such it is complementary to ongoing research on fake news detection \citep{Zellers2019Grover}, toxicity detection \citep{Han2020Fortifying, pavlopoulos2020toxicity}, and
consistency detection in, e.g., summarization \citep{Maynez2020OnFA, wang2020asking}. 
Human subjects (crowdworkers) were paid at a rate higher the legal minimum wage in our locale (Washington State, U.S.A.).

\bibliographystyle{acl_natbib}
\bibliography{acl2021}

\end{document}